\documentclass{article} 
\PassOptionsToPackage{sort}{natbib}

\usepackage{iclr2022_conference,times}

\usepackage{url}
\usepackage{amsmath}

\usepackage{amssymb}
\usepackage{mathtools}
\usepackage{mathrsfs}
\usepackage{upgreek}
\usepackage{mleftright}
\usepackage[pdftex,pdfstartview=FitB]{hyperref}
\usepackage{esint}
\usepackage{bm}
\usepackage{bbm}
\usepackage{bbold}
\usepackage{url}            
\usepackage{booktabs}       
\usepackage{amsfonts}       
\usepackage{nicefrac}       
\usepackage{microtype}      
\usepackage{mathrsfs}
\usepackage{color}
\usepackage{tabu}
\usepackage{multirow}
\usepackage{float}
\usepackage{booktabs}
\usepackage[ruled,vlined]{algorithm2e}
\usepackage{color}
\usepackage{tabu}
\usepackage{comment}
\usepackage[standard,hyperref,amsmath]{ntheorem}
\usepackage{graphicx}
\usepackage{epstopdf}
\usepackage{subcaption}
\usepackage{framed}
\usepackage{enumitem}
\usepackage{textcomp}
\usepackage{setspace}
\usepackage{cleveref}
\usepackage{latexsym}
\usepackage{alltt}
\usepackage{stmaryrd}
\usepackage{xspace}
\usepackage{euscript}
\usepackage{xfrac}    
\usepackage{nicefrac} 
\usepackage{tikz}
\usepackage{tikz-qtree}
\usepackage[normalem]{ulem}

\definecolor{mydarkblue}{rgb}{0,0.08,0.45}
\hypersetup{ %
	colorlinks=true,
	linkcolor=mydarkblue,
	citecolor=mydarkblue,
	filecolor=mydarkblue,
	urlcolor=mydarkblue}

\newcommand{\scrL}{\mathscr{L}}

\newcommand{\scrP}{\mathscr{P}}

\newcommand{\euA}{\EuScript{A}}

\newcommand{\euD}{\EuScript{D}}
\newcommand{\euE}{\EuScript{E}}

\newcommand{\euG}{\EuScript{G}}

\newcommand{\euN}{\EuScript{N}}
\newcommand{\euO}{\EuScript{O}}

\newcommand{\euS}{\EuScript{S}}

\newcommand{\euV}{\EuScript{V}}

\newcommand{\Ex}{\mathbb{E}}

\newcommand{\RR}{\mathbb{R}}

\DeclareMathOperator*{\Argmax}{Argmax}

\newcommand{\bpi}{\bm{\pi}}

\newcommand{\btheta}{\bm{\theta}}

\newcommand{\bphi}{\bm{\phi}}

\newcommand{\sfA}{\mathsf{A}}

\newcommand{\sfN}{\mathsf{N}}

\renewcommand{\mid}{\,|\,}
\newcommand{\midd}{\,|\kern-0.25ex|\,}

\DeclareMathAlphabet\rsfscr{U}{rsfso}{m}{n}

\crefname{assumption}{Assumption}{Assumptions}
\Crefname{assumption}{Assumption}{Assumptions}

\crefname{problem}{Problem}{Problems}
\Crefname{problem}{Problem}{Problems}

\makeatletter
\DeclareFontFamily{OMX}{MnSymbolE}{}
\DeclareSymbolFont{MnLargeSymbols}{OMX}{MnSymbolE}{m}{n}
\SetSymbolFont{MnLargeSymbols}{bold}{OMX}{MnSymbolE}{b}{n}
\DeclareFontShape{OMX}{MnSymbolE}{m}{n}{
	<-6>  MnSymbolE5
	<6-7>  MnSymbolE6
	<7-8>  MnSymbolE7
	<8-9>  MnSymbolE8
	<9-10> MnSymbolE9
	<10-12> MnSymbolE10
	<12->   MnSymbolE12
}{}
\DeclareFontShape{OMX}{MnSymbolE}{b}{n}{
	<-6>  MnSymbolE-Bold5
	<6-7>  MnSymbolE-Bold6
	<7-8>  MnSymbolE-Bold7
	<8-9>  MnSymbolE-Bold8
	<9-10> MnSymbolE-Bold9
	<10-12> MnSymbolE-Bold10
	<12->   MnSymbolE-Bold12
}{}

\let\llangle\@undefined
\let\rrangle\@undefined
\DeclareMathDelimiter{\llangle}{\mathopen}%
{MnLargeSymbols}{'164}{MnLargeSymbols}{'164}
\DeclareMathDelimiter{\rrangle}{\mathclose}%
{MnLargeSymbols}{'171}{MnLargeSymbols}{'171}
\makeatother

\newcommand{\euclidnorm}[1]{\left\lVert#1\right\rVert_2}

\newcommand{\matsnorm}[2]{|\kern-0.25ex|\kern-0.25ex| #1 |\kern-0.25ex|\kern-0.25ex|_{{#2}}}

\renewcommand{\left}{\mleft}
\renewcommand{\right}{\mright}

\title{The Multi-Agent Pickup and Delivery Problem: MAPF, MARL and Its Warehouse Applications
}

\author{Tim Tsz-Kit Lau
	\\
	Department of Statistics\\
	Northwestern University\\
	\texttt{timlautk@u.northwestern.edu} \\
	\And
	Biswa Sengupta \\
	Zebra Technologies \\
	Endeavour House, Shaftsbury Avenue, London, UK \\
	\texttt{biswa.sengupta@zebra.com}
}

\iclrfinalcopy 
\begin{document}

	\maketitle
	\begin{abstract}
		We study two state-of-the-art solutions to the multi-agent pickup and delivery (MAPD) problem based on different principles---multi-agent path-finding (MAPF) and multi-agent reinforcement learning (MARL). Specifically, a recent MAPF algorithm called conflict-based search (CBS) and a current MARL algorithm called shared experience actor-critic (SEAC) are studied. While the performance of these algorithms is measured using quite different metrics in their separate lines of work, we aim to benchmark these two methods comprehensively in a simulated warehouse automation environment.

	\end{abstract}

	\section{Introduction}
	\label{sec:intro}
	The multi-agent pickup and delivery (MAPD) problem in various industrial applications such as warehouse automation has long been a central problem to study in artificial intelligence due to its widespread real-world applications. In the MAPD problem, various agents operate in an everyday environment in a multi-agent system. Each of them picks up a new item in the request queue and delivers it to a designated delivery location. To execute these tasks, the agents need to travel in the environment via their collision-free paths. To be more precise, each agent has to move from its current location to the pickup location of a requested item in the request queue and travel to the delivery location of the item after the pickup. 
	
	Two major approaches to tackle MAPD are multi-agent path-finding (MAPF) and multi-agent reinforcement learning (MARL). 	
	MAPF, as a more traditional solution to the MAPD problem, involves computing collision-free paths for many agents given the current states of the agents (e.g., their locations) and a representation of the environment (e.g., the pickup and delivery locations and obstacles present in the environment). MAPF also finds applications in computer games, traffic management and airport schedules. 
	Most methods for solving the MAPD problem in the literature are based on MAPF, so somehow, MAPD and MAPF are viewed as the same problem. 
	A noticeable difference between them is that a vanilla MAPF problem assumes the planning problem is \emph{single-shot}, i.e., the agents will stay at their goal locations once they have arrived. In contrast, an MAPD problem is usually \emph{lifelong}, i.e., an agent will begin delivery once after a pickup and will travel to a new requested item location once it has finished a delivery. This paper aims to tackle the MAPD problem with the more realistic lifelong MAPF approaches.

	Due to the recent interest in MARL, the MAPD problem is also solved with MARL algorithms. However, it is generally viewed as a benchmark problem to showcase the efficiency of newly proposed MARL algorithms instead of a specific problem to study. Based on the formulation of the MARL problems, MARL algorithms tackle the MAPD problem in a completely different flavour. They do not compute a full collision-free path but instead learn agents' policies that choose actions of the agents given their current observations of the environment. These agents' policies are learned to maximize the agents' cumulative (discounted) rewards, which are usually assigned when the agents have successfully picked up or delivered an item. 
	
	The qualities of the above two types of MAPD solutions are usually assessed using very different metrics in their lines of work. The MAPF-based solutions are mainly evaluated using success rates, flow times (the sum of arrival times of all agents at their goal locations)
	and makespans (the maximum of the arrival times of all agents at their goal locations). In contrast, the MARL-based methods are usually assessed with the standard metrics used in reinforcement learning---training and evaluation returns. Such a discrepancy between the evaluation metrics renders it difficult in comparing these two types of solutions. Given this discrepancy, in this paper, we aim to provide a comprehensive comparison between these two seemingly disconnected types of solutions to the MAPD problem. We particularly compare the lifelong version of a well-known centralized single-shot MAPF solver called conflict-based search (CBS) to the state-of-the-art MARL solver called shared experience actor-critic (SEAC).

	\section{Background}
	In this section, we give a general overview of the problem formulations of MAPF and MARL, which can also be found from prior work \citep{li2019disjoint,huang2021learning,li2019improved,li2021lifelong,liu2019task,christianos2020shared,christianos2021scaling}. Details of related algorithms are given in \Cref{sec:algo}.

	\subsection{Single-shot Multi-Agent Path Finding} 
	\label{subsec:mapf}
	A (single-shot) multi-agent pathfinding problem is defined by an unweighted undirected graph $\euG = (\euV, \euE)$ and a set of $n$ agents $\sfA \coloneqq\{\alpha_1, \ldots, \alpha_n\}$. Each agent $\alpha_i$ has a start vertex $s_i\in\euV$ and a goal vertex $g_i\in\euV$. With time discretized into time steps, each agent can only either move to an adjacent vertex or stay at the current vertex in the graph at each time step. Both of the move and wait actions incur a unit cost until the agent has arrived at its goal vertex and no longer moves so that the cost of each agent is the number of time steps required for its arrival at its goal vertex from its start vertex. There are two types of conflicts under consideration: (i) a \emph{vertex conflict}, denoted by $\langle \alpha_i, \alpha_j, v, t\rangle$, happens when agents $\alpha_i$ and $\alpha_j$ are at the same vertex $v\in\euV$ at time step $t$; (ii) an \emph{edge conflict}, denoted by $\langle \alpha_i, \alpha_j, v_1, v_2, t\rangle$, occurs when agents $\alpha_i$ and $\alpha_j$ traverse the same edge $(v_1,v_2)\in\euE$ in opposite directions between time steps $t$ and $t+1$. The overall objective of MAPF is to find a set of conflict-free paths which move all agents from their start vertices to their goal vertices, which are often referred to as \emph{solutions}, by minimizing the sum of the costs of all the agents.

	\subsection{Lifelong Multi-Agent Path Finding}
	As mentioned in \Cref{sec:intro}, an MAPD problem is indeed a lifelong MAPF problem that solves possibly multiple single-shot MAPF problems in an inner loop. Thus, existing approaches for solving lifelong MAPF problems are usually based on those for solving single-shot MAPF instances, which can be categorized into three main types. 
	\begin{enumerate}
		\item A lifelong MAPF problem is decomposed into a sequence of single-shot MAPF instances where all agents perform path replanning at every step. 
		\item A lifelong MAPF problem is decomposed into a sequence of single-shot MAPF instances where path replanning is performed only for agents that have just picked up or delivered their items \citep{ma2017lifelong}. 
		\item A lifelong MAPF problem is solved as a whole in an offline setting, as reductions to other well-studied combinatorial problems such as an answer set programming problem \citep{nguyen2017generalized}.
	\end{enumerate}
	See \citet{li2021lifelong} for more detailed descriptions of these three types of solutions to lifelong MAPF problems, \citet{ma2017overview,felner2017search} for detailed surveys on MAPF (mainly single-shot), and also \citet{salzman2020research} for recent research challenges and opportunities in MAPF and MAPD problems.

	\subsection{Multi-Agent Reinforcement Learning}
	Solutions based on multi-agent reinforcement learning (MARL), unlike MAPF-based solutions, do not compute a set full collision-free paths of all the agents in the environment, but instead learn policies of the agents which decide the actions of the agents at each time step given the current states of the environment. 
	
	This MARL problem can be formulated as a \emph{partially observable} multi-agent Markov decision process (a.k.a.~Markov game) for $n$ agents, which is defined by the tuple $(\euN, \euS, \{\euO_i\}_{i\in\euN}, \{\euA_i\}_{i\in\euN}, P, \{R_i\}_{i\in\euN})$, where $\euN\coloneqq\{1,\ldots, n\}$ denotes the set of $n$ agents, $\euS$ is the state space, $\euO\coloneqq\euO_1\times\cdots\times\euO_n$ is the joint observation space, $\euA\coloneqq\euA_1\times\cdots\times\euA_n$ is the joint action space. Each agent $i$ can only perceive local observations $o_i\in\euO_i$ which depend on the current state. The function $P\colon\euS\times\euA\to\scrP(\euS)$, which is known as a transition model, returns a distribution on the successive state given the current state and joint action. For each agent $i$, the reward function $R_i\colon\euS\times\euA\times\euS\to\RR$ gives its individual reward $r_{i,t}$ at time step $t$. The overall MARL objective is to find an optimal joint policies of the agents, denoted by $\bpi^\star=(\pi_1^\star, \ldots, \pi_n^\star)$, such that the discounted return of each agent $i$ is maximized with respect to the policies of other agents, i.e., 	
	\begin{equation}\label{eqn:mdp_obj}
		(\forall i\in\euN)\quad \pi_i^\star \in \Argmax_{\pi_i} \ \Ex_{\pi_i, \pi_{\setminus i}^\star}\left[\sum_{t=0}^{T} \gamma^t r_{i,t}\right],
	\end{equation}
	where $\pi_{\setminus i} \coloneqq (\pi_1, \ldots, \pi_{i-1}, \pi_{i+1}, \ldots, \pi_n)$, $\gamma\in(0,1]$ is the discount factor, and $T$ is the total number of time steps of an episode. 
	
	Note that, based on the different contexts of MARL algorithms, we can add more stringent assumptions to the above formulation, e.g., the action spaces, the observation spaces, or the reward functions of the agents can be assumed to be identical \citep{christianos2020shared,rashid2018qmix,foerster2018counterfactual}.

	\section{Algorithms for MAPF and MARL Problems}
	\label{sec:algo}
	In this section, we give the details of several popular algorithms for MAPF and MARL problems which are compared numerically in \Cref{sec:expt}.

	\subsection{MAPF Solvers}
	\subsubsection{Conflict-Based Search for Single-shot MAPF}
	While there are a multitude of MAPF solvers developed in recent years, conflict-based search (CBS) and its variants are among the strongest algorithms. 
	Conflict-based search \citep[CBS;][]{sharon2015conflict} is a centralized bilevel tree search algorithm, which resolves conflicts by adding constraints at the high level and replans paths for agents respecting these constraints at the low level. At the high level, CBS performs a best-first search on the \emph{constraint tree} (CT), which is a binary search tree, according to the costs of the CT nodes. Each CT node $\sfN$ encompasses: 	
	\begin{enumerate}
		\item a set of constraints $\sfN_\text{constraints}$ in the search, in which each constraint can be either a vertex constraint or an edge constraint (see \Cref{subsec:mapf} for their definitions); 
		
		\item a solution $\sfN_\text{solution}$ which consists of a set of individually cost-minimal paths for all agents, subject to the constraints in $\sfN_\text{constraints}$; 
			
		\item a cost $\sfN_\text{cost}$ of $\sfN$ which is the sum of costs of the paths in $\sfN_\text{solution}$;
	
		\item a set of conflicts $\sfN_\text{conflicts}$ between any two paths in $\sfN_\text{solution}$. 	
	\end{enumerate}
	At the high level, CBS begins with only one node with an empty constraint set and expands the CT by expanding a CT node with the lowest cost $\sfN_\text{cost}$. After choosing such a CT node to expand, CBS finds the set of conflicts $\sfN_\text{conflicts}$ in $\sfN_\text{solution}$. If there are none, CBS terminates and returns $\sfN_\text{solution}$. Otherwise, CBS \emph{randomly} chooses one of the conflicts to resolve by splitting $\sfN$ into two child CT nodes. In each of these two children, we add CT nodes, an additional vertex or edge constraint on one of the two conflicting agents to $\sfN_\text{constraints}$ of the corresponding child node, depending on the type of the conflict. This is done similarly for the other conflicting agent and its corresponding child node. 
	
	At the low level, after the addition of the constraints, path replanning is performed in $\sfN_\text{solution}$ whenever necessary via a low-level search such as cooperative A* search \citep{silver2005cooperative}, while keeping other paths unchanged. A child CT node will be pruned if this low-level search cannot find any path that satisfies the constraints.

	Improved versions of CBS have also been developed to improve its efficiency. Improved CBS \citep[ICBS;][]{boyarski2015icbs} prioritizes the conflicts to split on at each CT node $\sfN$, whereas CBSH \citep{felner2018adding} accelerates the high-level search through an additional admissible heuristic instead of always choosing to expand the CT node with the lowest cost $\sfN_\text{cost}$. 	
	See also e.g., \citet{huang2021learning,li2019improved,li2019disjoint,li2021lifelong,barer2014suboptimal,ma2019searching} for more details and recent advances of CBS and its variants for MAPF.

	While the above CBS algorithms are mainly for single-shot MAPF problems, a common approach to solve lifelong MAPF is to stitch a sequence of single-shot MAPF instances together by using a MAPF solver to replan whenever at least one agent is assigned to a new target location at each time step, see e.g., \cite{liu2019task}. Since replanning time grows exponentially with the number of agents, reducing replanning frequencies such as planning paths within a finite window \citep{li2021lifelong} improves the scalability of lifelong MAPF.

	\subsection{MARL Algorithms}	
	\subsubsection{Policy Gradient and Actor-Critic Algorithms}
	The policy gradient algorithm such as REINFORCE \citep{williams1992simple} is a model-free reinforcement learning algorithm which learns an optimal policy $\pi_{\btheta}$, usually parameterized by $\btheta$. The expected return is defined through $J(\btheta) \coloneqq\Ex_{s\sim \euD^\pi, a\sim\pi_{\btheta}(\cdot\mid s)}[Q^{\pi_{\btheta}}(s, a)]$, with
	\[
	Q^{\pi_{\btheta}}(s,a) \coloneqq \Ex_{a\sim\pi_{\btheta}(\cdot\mid s)}\left[\left.\sum_{k=0}^T \gamma^k r_{t+k+1} \,\right|\, s_t =s, a_t=a\right],
	\]
	where $\euD^\pi$ is the on-policy state distribution under $\pi$. 	
	To maximize the expected return $J$, we compute the gradient of the objective via the policy gradient theorem \citep{sutton2000policy}, which gives  
	\[
	\nabla_{\btheta}J(\btheta) = \Ex_{s\sim \euD^\pi, a\sim\pi_{\btheta}(\cdot\mid s)}\left[Q^{\pi_{\btheta}}(s, a) \nabla_{\btheta} \log \pi_{\btheta}(s, a)\right]. 
	\]
	While the Markov property is not used in computing policy gradients, so that they can be used in partially observable settings,  the estimation of policy gradients often suffer from high variance. To achieve variance reduction, actor-critic algorithms estimate Monte Carlo returns with a value function $V_{\bphi}^\pi(s)$ parameterized by $\bphi$. Hence, an actor-critic algorithm under a multi-agent partially observable setting makes use of the following policy loss for agent $i$ 	
	\begin{equation*}\label{eqn:policy_loss}
		\scrL(\btheta_i) \coloneqq -\log\pi_{\btheta_i}(a_t^i\mid o_t^i)\cdot\left[r_t^i + \gamma V_{\bphi_i}(o_{t+1}^i) -  V_{\bphi_i}(o_{t}^i) \right], 
	\end{equation*}
	where the value function $V_{\bphi_i}$ minimizes
	\begin{equation*}\label{eqn:value_loss}
		\scrL(\bphi_i) \coloneqq \euclidnorm{V_{\bphi_i}(o_{t}^i) - y_i^{\bphi_i}}^2 \quad\text{with}\quad y_i^{\bphi_i} \coloneqq r_t^i + \gamma V_{\bphi_i}(o_{t+1}^i). 
	\end{equation*}
	In an implementation, both the policies and value functions are parameterized by neural networks. A2C \citep{mnih2016asynchronous} is used with $n$-step rewards, parallel trajectory sampling and entropy regularization. 
	
	\subsubsection{Shared Experience Actor-Critic}
	Based on the actor-critic algorithms described above, shared experience actor-critic \citep[SEAC;][]{christianos2020shared} is proposed for efficient learning using shared experience among agents. The main merit of sharing experience among agents is that agents can learn from the experiences of other agents without having the same rewards. In SEAC, the trajectories of other agents are used as off-policy data, and importance sampling with a behavioural policy $\rho$ is used to correct for the off-policy data. Detailed algorithmic descriptions of SEAC can be found in \citet{christianos2020shared}.

	\subsubsection{Other MARL Algorithms}
	As the MAPD problem involves cooperation among agents, a popular paradigm in such a cooperative MARL problem is the Centralized Training with Decentralized Execution. All agents can access data from all other agents during training but not at execution time. In addition to SEAC implemented in this paper, MARL algorithms of this type also include MADDPG \citep{lowe2017multi}, Q-MIX \citep{rashid2018qmix} and COMA \citep{foerster2018counterfactual}. Added to the above general MARL algorithms, two recent works by \citet{sartoretti2019primal} and \citet{damani2021primal} develop specific MARL algorithms to the lifelong MAPF problem.

	\section{Numerical Experiments}
	\label{sec:expt}
	
	We evaluate the MAPF-based and MARL-based methods on a simulated robotic warehouse environment, which is modified from 
	the Multi-Robot Warehouse Environment \citep[RWARE;][]{papoudakis2021benchmarking}. The modifications are mainly for more efficient training of SEAC and the applicability of the lifelong version of CBS. First, agents need to pick up and deliver requested shelves (items) to the delivery locations but do not need to return the shelves to empty shelves before travelling to new pickup locations. 
	Furthermore, the number of delivery locations is increased so that agents can deliver items in any location of the bottom row. This modification is done because each agent should have its distinct goal location in CBS. 
	Finally, to address the issue of sparse rewards when training MARL algorithms, each agent will be assigned a $+1$ reward when picking up a requested item and a $+2$ reward when delivering items to a delivery location successfully. Other specifications such as the observations, actions and dynamics remain the same as in RWARE. 
	
\begin{figure}[h]
		\centering
		\begin{subfigure}[b]{0.3\textwidth}
			\centering
			\includegraphics[width=0.75\textwidth]{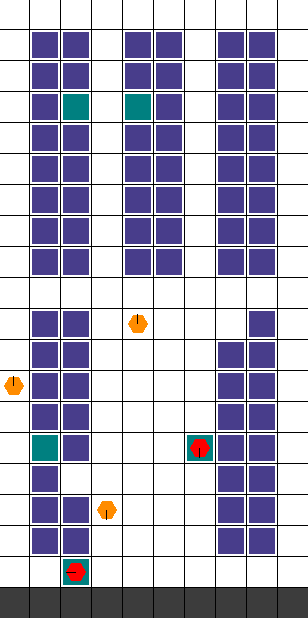}
			\caption{small; 5 agents}
			\label{fig:small}
		\end{subfigure}
		\hspace*{3.5mm}
		\begin{subfigure}[b]{0.3\textwidth}
			\centering
			\includegraphics[width=\textwidth]{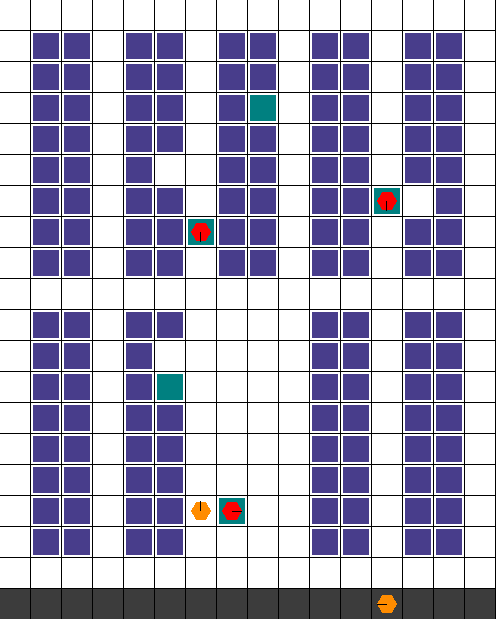}
			\caption{medium; 5 agents}
			\label{fig:medium}
		\end{subfigure}
		\hfill
		\begin{subfigure}[b]{0.3\textwidth}
			\centering
			\includegraphics[width=0.8\textwidth]{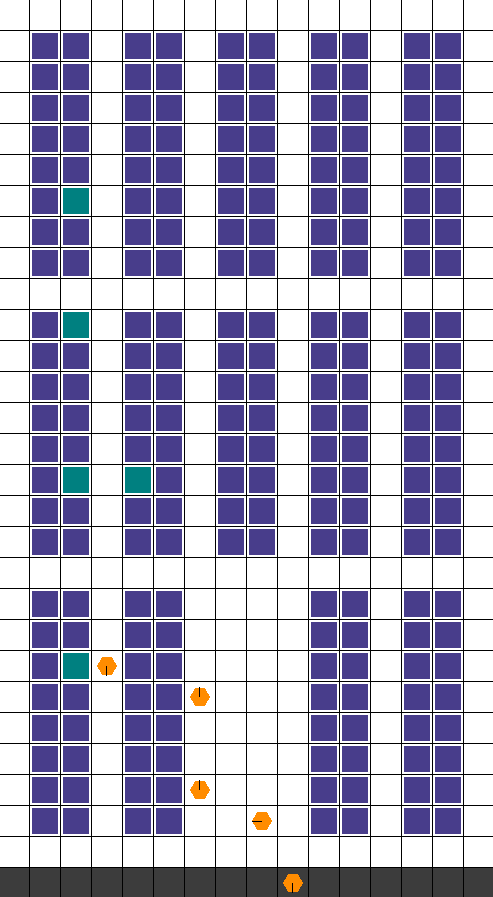}
			\caption{large; 5 agents}
			\label{fig:large}
		\end{subfigure}
		\caption{The modified RWARE environment of different sizes.}
		\label{fig:rware}
	\end{figure}

	The modified robotic warehouse environment is visualized in \Cref{fig:rware}, in three different sizes (small, medium and large). Each agent is hexagonal, with a black line indicating its facing direction. When an agent is not carrying any item in yellow, its target is to move to a requested item that is green then pick it up. An agent carrying an ordered item is indicated by red, and its target is to deliver the item to any location of the grey bottom row. A new item will be randomly generated in the request queue right after the delivery of an object so that the number of items in the request queue remains unchanged. Each agent repeats this pickup and delivery cycle until an episode of a fixed number of time steps ends. 
	When an agent is not carrying any item, it can move through any coordinates in the environment, including under the unrequested purple shelves. It can only move across the corridors and the delivery row if it carries an item. 
	We also assumed an equal number of agents and requested items in the environment to simplify the tasks.
	
	Regarding the MAPF-based method, we have implemented a centralized single-shot MAPF algorithm, the vanilla single-shot CBS algorithm, and solved the lifelong MAPF problem by replanning whenever any agent has a new goal location. This solution would imply that replanning becomes more frequent whenever the number of agents increases or the environment becomes denser. On the other hand, as a MARL-based method for comparison purposes, we train a SEAC algorithm based on this modified environment with a less sparse reward design. Empirically we observe that this modification leads to lower sample complexity than the original one. All the experiments are run on a machine with an Intel Xeon E5-2699v4 CPU, a single Tesla V100 GPU (only used in SEAC), and 540GB RAM.
	
	One remarkable difference between the implementations of lifelong CBS and SEAC is that the vanilla CBS is homogeneous among all agents so that all items except the requested ones are viewed as obstacles in the environment. This implies all agents, regardless of the status of carrying items or not, cannot move under any unrequested shelves. This significantly reduces the number of feasible paths in the lifelong CBS solver. Therefore, it is expected that the lifelong CBS solver will fail whenever the environment is too dense, i.e., the number of agents relative to the size of the domain is too large.

	We compare lifelong CBS and SEAC at test time, considering all three sizes of the warehouse, with 2, 5 and 8 agents in lifelong CBS and 5, 10 and 15 agents in SEAC. We observe that lifelong CBS does not scale to 10 or more agents in any of the sizes of the environment due to time-consuming replanning schemes. We evaluate their performance with five random seeds, each with four episodes of 500-time steps. These two algorithms are compared using the following five metrics (averaged over all episodes and random seeds), which are commonly used in either MAPF or MARL but not both: 	
	(i) mean flow time for the first delivery (the sum of arrival times of all agents at their delivery locations, in terms of the number of time steps required); 
	(ii) mean makespan for the first delivery (the maximum of the arrival times of all agents at their delivery locations, in terms of the number of time steps required); 
	(iii) mean episodic cumulative reward per agent in one episode; 
	(iv) mean number of successfully delivered items of each agent in one episode; 
	(v) mean episodic time (in seconds). 
	These metrics for lifelong CBS and SEAC are given in \Cref{table:metrics_cbs,table:metrics_seac} respectively. 	
	\begin{table}[h]
		\centering
		\footnotesize
		\caption{Metrics (s.e.) for lifelong CBS. }
		\label{table:metrics_cbs}
		\vspace*{1mm}
		\begin{tabular}{crrrr}
			\toprule
			&& \multicolumn{3}{c}{Lifelong CBS} \\
			\cmidrule[0.5pt]{3-5}
			Metric & $n$ & small & medium & large   \\
			\midrule
			\multirow{3}{*}{\textbf{\shortstack{Mean\\ flowtime}}} & 2 & 66.60 (7.58) & 75.20 (17.68)  &  76.00 (17.13)  \\
			& 5 & 178.60 (16.40) & 196.20 (29.10) & 209.80 (11.38)   \\			
			& 8 & 307.00 (53.43) & 345.60 (37.71) & 339.80 (22.38)   \\
			\midrule
			\multirow{3}{*}{\textbf{\shortstack{Mean\\ makespan}}} & 2  & 41.40 (5.30) & 44.60 (10.29) & 44.80 (11.08) \\
			& 5 & 51.60 (5.01)  & 55.40 (9.30) &  60.40 (7.33)   \\
			& 8  & 61.60 (8.50) & 71.80 (9.89) & 66.20 (4.07)  \\
			\midrule	
			\multirow{3}{*}{\textbf{\shortstack{Mean\\ cumulative\\ reward}}} & 2  & 36.02 (2.98) & 35.30 (2.09) & 27.95 (1.89)  \\
			& 5 & 31.19 (1.24)  & 30.63 (4.59) &  26.09 (1.36)  \\
			& 8  & 27.19 (1.39) & 28.02 (1.19) &  23.40 (1.16)  \\			
			\midrule
			\multirow{3}{*}{\textbf{\shortstack{Mean\\ \# delivered\\ items}}} & 2  & 30.30 (14.22) & 30.20 (13.83) & 23.90 (11.01)  \\
			& 5 & 27.07 (12.63)  & 27.87 (15.10) & 22.48 (10.43)  \\
			& 8  & 24.50 (11.61) & 24.46 (11.29) & 20.38 (9.44) \\			
			\midrule
			\multirow{3}{*}{\textbf{\shortstack{Mean\\ episodic\\ time}}} & 2  & 508.40 (0.97) & 510.57 (0.78) & 513.53 (1.26)  \\
			& 5 & 567.34 (79.62)  & 637.12 (340.26)  &  565.88 (29.60)  \\
			& 8  & 2618.11 (4866.68) & 1391.95 (1372.68) &  2371.99 (2409.23) \\			
			\bottomrule
		\end{tabular}
	\end{table}

	\begin{table}[h]
		\centering
		\footnotesize
		\caption{Metrics (s.e.) for SEAC. }
		\label{table:metrics_seac}
		\vspace*{1mm}
		\begin{tabular}{crrrr}
			\toprule
			&& \multicolumn{3}{c}{SEAC}\\
			\cmidrule[0.5pt]{3-5}
			Metric & $n$ & small & medium & large  \\
			\midrule
			\multirow{3}{*}{\textbf{\shortstack{Mean\\ flowtime}}} & 5 & 274.15 (53.04) & 367.00 (137.32) & 555.90 (182.28) \\
			& 10 & 482.55 (88.30) & 763.90 (223.34) & 805.45 (146.45) \\
			& 15 & 882.15 (201.03) & 846.20 (127.07) & 1030.80 (268.63) \\
			\midrule
			\multirow{3}{*}{\textbf{\shortstack{Mean\\ makespan}}} & 5  &  88.50 (27.37)  & 142.50 (65.43) & 221.70 (93.63) \\
			& 10  & 94.80 (33.62) & 203.65 (101.51) & 177.85 (58.99) \\
			& 15  & 185.75 (90.75) & 146.45 (44.56) & 305.05 (110.42)  \\
			\midrule	
			\multirow{3}{*}{\textbf{\shortstack{Mean\\ cumulative\\ reward}}} & 5 &  37.01 (2.53)  & 25.98 (5.04) & 12.26 (3.06) \\
			& 10  & 39.60 (2.01) & 21.37 (4.03) & 19.42 (4.47) \\
			& 15  & 31.02 (3.93) & 23.29 (4.08) & 8.18 (1.17) \\
			\midrule
			\multirow{3}{*}{\textbf{\shortstack{Mean\\ \# delivered\\ items}}} & 5 & 10.00 (9.69)  & 14.20 (33.69) & 5.71 (14.21) \\
			& 10 & 8.41 (5.72) & 5.12 (7.84) & 8.69 (22.86) \\
			& 15 & 12.90 (19.53) & 16.77 (34.98) & 30.77 (59.61) \\
			\midrule
			\multirow{3}{*}{\textbf{\shortstack{Mean\\ episodic\\ time}}} & 5 & 509.75 (0.23)  & 511.34 (0.39) & 513.50 (0.40) \\
			& 10 & 512.77 (0.36) & 514.43 (0.42) & 516.66 (0.45) \\
			& 15 & 515.67 (0.50) & 517.59 (0.43) & 519.48 (0.46) \\
			\bottomrule
		\end{tabular}
	\end{table}
	
	From \Cref{table:metrics_cbs}, we observe that, in general, both the mean flow time and makespan increase with the number of agents and the size of the environment. The increased environment size leads to a less dense environment, so more time steps are required for completing deliveries due to longer travel distances. On the other hand, the increase in the number of agents gives rise to a denser environment so that there are fewer possible collision-free paths. Agents need to travel long distances to avoid collisions which are more likely to occur in a denser environment. 	
	In addition, the mean cumulative reward and the mean number of delivered items of each agent per episode decrease with the number of agents and the size of the environment. Within a fixed time of 500 in every episode, each agent can only deliver fewer items in an environment with more agents or of larger size, thus receiving a smaller reward. 
	
	More notably, while the mean episodic times of lifelong CBS are close in different sizes of the environment, the mean episodic time grows significantly with the number of agents, also suffering from high variance. Since our implementation of lifelong CBS is online, episodic times include the time required for both replanning and agents' movement in the environment. Attributed to a denser environment, replanning is performed at higher frequencies with more agents. Each replanning is also more time-consuming as more conflicts have to be resolved. The high variance of the mean episodic times with large numbers of agents also indicates that lifelong CBS's performance is not robust enough in different instances of the environment. 
	
	For SEAC, we observe from \Cref{table:metrics_seac} similar variations of the mean flow time and the mean makespan. However, with more agents in the environment, agents trained using SEAC can deliver more items with more agents on average in an episode. This is a remarkable difference from lifelong CBS, which has significantly fewer possible paths in the warehouse than SEAC. 
	
	Furthermore, since the agents' policies in SEAC are learned during training and the mean episodic times are computed at test time, only agents' environmental movements account for the episodic time. Therefore, the mean episodic times are very close regardless of the size of the environment or the number of agents. This makes SEAC a more viable and efficient solution in practice as robots have to stay idle during replanning in lifelong CBS.

	Comparing lifelong CBS and SEAC from \Cref{table:metrics_cbs,table:metrics_seac} for 5 agents, spending similar time in each episode, lifelong CBS has significantly smaller mean flowtime and mean makespan, while having significantly more items delivered in an episode by each agent on average. This is because CBS-based methods plan the shortest collision-free path for every individual agent from its start location to its target location, which should be more efficient than SEAC which only gives the most probable action for each agent based on the current observation of the environment at every time step (i.e., no planning ahead). However, depending on the density of the environment, MAPF-based methods suffer from the issue of scalability. Replanning in lifelong CBS might take a very long time in practice or even fail in various instances in a relatively dense environment. This suggests that lifelong CBS is a more efficient solver for the MAPD problem with fewer agents and in less dense environments, whereas SEAC should be used when many agents are used. However, we should note that the training of SEAC agents also costs a significant amount of time, which mainly rises with the number of agents.

	\section{Discussion and Future Work}
	While we implemented the lifelong CBS with the principle that replanning is performed whenever any one of the agents changes its target locations, this current implementation is far from efficient. In addition to the improved variants of CBS mentioned in \Cref{sec:algo}, various recent works in MAPF address different perspectives and improve the efficiencies of MAPF solvers, see e.g., \citet{wu2021multi,shahar2021safe,greshler2021cooperative,honig2019persistent,ma2019lifelong,huang2021learning}.
	
	Furthermore, a crucial and valuable research direction to pursue is to use lifelong MAPF methods to improve the sample efficiency of MARL algorithms for solving the MAPD problem. For instance, the solutions based on lifelong MAPF solvers can be used as expert demonstration data to derive a policy, i.e., (multi-agent) imitation learning \citep{ho2016generative,song2018multi,wang2021multi,lin2021decentralized}. Furthermore, when the reward functions are hard to design or each agent has no access to the rewards or goals of other agents, we can use expert demonstration data to learn the rewards, i.e., (multi-agent) inverse reinforcement learning \citep{yu2019multi,filos2021psiphi}. 
	
	The MAPD problem considered in this paper is a relatively simple instance where all agents are assumed to be homogeneous: all agents can pick up any requested item and deliver items at any delivery location. Instead of the implicit assumption of all agents being homogeneous, a more realistic yet complicated scenario is to allow agents to have different abilities and goals: each agent can only pick up a designated type of item. A recent MARL algorithm based on selective parameter sharing \citep[SePS;][]{christianos2021scaling} is designed to handle such settings, in which parameter sharing is performed among individual groups of homogeneous agents with grouping performed automatically using an unsupervised clustering algorithm based on the abilities and goals of the agents. It remains to see how SePS compares to lifelong MAPF-based solutions in more complicated MAPD problems.

	\newpage
	

	\bibliography{ref}
	\bibliographystyle{iclr2022_conference}
	
	\newpage
	\appendix
	\numberwithin{equation}{section}
	\numberwithin{theorem}{section}
	\numberwithin{proposition}{section}
	\numberwithin{lemma}{section}
	\numberwithin{definition}{section}
	\numberwithin{corollary}{section}
	\numberwithin{example}{section}
	\numberwithin{remark}{section}
	\numberwithin{problem}{section}

\end{document}